\documentclass[twocolumn]{article}
\usepackage{graphicx} 
\usepackage{multirow}
\usepackage{bm} 
\usepackage{enumitem}
\usepackage{lipsum}  
\usepackage[english]{babel}
\usepackage[letterpaper,top=2cm,bottom=2cm,left=2cm,right=2cm,marginparwidth=1.75cm]{geometry}
\usepackage{amsmath} 
\usepackage{amsmath}
\usepackage{graphicx}
\usepackage[colorlinks=true, allcolors=blue]{hyperref}
\usepackage{titling}
\usepackage{hyperref}
\usepackage{caption}
\usepackage{cite}
\usepackage{xcolor}

\title{WMINet: A Wheel-Mounted Inertial Learning Approach For Mobile-Robot Positioning
}
\author{Gal Versano, Itzik Klein\\
The Hatter Department of Marine Technologies,\\
Charney School of Marine Sciences, University of Haifa, Haifa,\\
Israel}
\date{}
\begin{document}
\maketitle


\begin{abstract}
Autonomous mobile robots are widely used for navigation, transportation, and inspection tasks indoors and outdoors. In practical situations of limited satellite signals or poor lighting conditions, navigation depends only on inertial sensors. In such cases, the navigation solution rapidly drifts due to inertial measurement errors. In this work, we propose WMINet a wheel-mounted inertial deep learning approach to estimate the mobile robot's position based only on its inertial sensors. To that end, we merge two common practical methods to reduce inertial drift: a wheel-mounted approach and driving the mobile robot in periodic trajectories. Additionally, we enforce a wheelbase constraint to further improve positioning performance. To evaluate our proposed approach we recorded using the Rosbot-XL a wheel-mounted initial dataset totaling 190 minutes, which is made publicly available. Our approach demonstrated a 66\% improvement over state-of-the-art approaches. As a consequence, our approach enables navigation in challenging environments and bridges the pure inertial gap. This enables seamless robot navigation using only inertial sensors for short periods.
\end{abstract}
\section{Introduction}
The increasing adoption of mobile robots is driven by key factors such as enhanced efficiency, improved productivity, and greater adaptability. Advancement in technology, the decreasing costs of electronic sensors and devices have led to a surge in research and demand for mobile robots \cite{borenstein1997mobile}, \cite{tzafestas2018mobile}. These robots are utilized in a wide range of industries, from agriculture—where they assist with tasks like fruit picking and crop monitoring—to logistics, enabling seamless indoor and outdoor deliveries. Additionally, they play a crucial role in construction, facilitating site inspections and material transportation. Mobile-robot rely on advanced sensor systems to autonomously perform tasks and navigate complex real-world environments \cite{desouza2002vision},  Examples include warehouse automation for inventory management, agricultural robots for precision farming, and autonomous vehicles for transportation and delivery \cite{item1}, \cite{item2}, \cite{item3}.\\
\noindent To complete their tasks, rather indoor or outdoor, the navigation system accuracy is a critical function. To meet the navigation requirements different sensors are used such as   vision-based systems \cite{item4}, \cite{item5}, global navigation satellite system receiver (GNSS) \cite{kaplan2017understanding}, and LiDAR \cite{item6},\cite{lan2024highly}. The latter can be used for scanning the environment and build a map so the robot can estimate its position \cite{item7}, \cite{wu2022wheel} in a process known as simultaneous localization and mapping (SLAM). As LiDARs are considered expansive devices, commonly cameras are employed for SLAM in indoors environments \cite{item8}, \cite{fu2024dynamic}. \\
In real-world scenarios poor lightening conditions, sharp maneuvers or GNSS outages often occur leading the navigation system to rely only on its inertial sensors in a process known as pure inertial navigation. There, the inertial sensor readings are used to calculate the navigation solution consisting of the position, velocity, and orientation \cite{farrell2008aided}. However, due to noise and errors in the sensor data, the navigation solution drifts over time. To reduce the inertial drift, the mobile robot pure inertial navigation (MoRPI) \cite{etzion2023morpi} was proposed. MoRPI requires the robot to move in periodic trajectories and uses an empirical formula to determine step length, similar to the approach used with quadrotors \cite{shurin2020qdr}. \\
In the above mentioned works, the inertial sensors are rigidly mounted on chassis of the vehicles. Recent works explore the possibility of mounted the inertial sensors on the mobile robot wheels. In \cite{9524467} and \cite{collin2014mems} a dead reckoning system for wheeled robots was proposed to highlight advantages of placing the inertial sensors at the center of a non-steering wheel. First, it can serve as an alternative to traditional odometers, helping to mitigate the error drift commonly associated with inertial navigation systems (INS). Second, as the wheel rotates, the constant bias error of the inertial sensor can be partially canceled. \\
Alongside advancements in wheel-mounted inertial navigation, deep learning algorithms have been increasingly employed to enhance the accuracy and robustness of inertial sensors and related fusion techniques \cite{COHEN2024103565}, \cite{aslan2022visual}, \cite{10492667}. With relevance to our work, MoRPINet \cite{etzion2024snakeinspiredmobilerobotpositioning} used a neural network to predict the traveled distance using only the inertial sensors, and calculates the heading angle by Madgwick Algorithm \cite{madgwick2010efficient}. \\
\noindent In this paper, we introduce WMINet an end-to-end deep-learning algorithm that predicts the displacement of mobile robot traveling in periodic trajectories using only wheel-mounted inertial sensors reading. Our key contributions are as follows:
\begin{enumerate}
    \item \textbf{WMINet} is a deep learning approach for estimating a mobile robot’s position using only wheel-mounted inertial sensors. 
    \item \textbf{Wheelbase Constraint}. Motivated by the information aided navigation concept \cite{engelsman2023information}, we leverage the known fixed distance between wheels into a dedicated loss function to improve the robot positioning.
    \item \textbf{Wheel-Mounted-IMU Dataset} We provide a dataset with a 38 minutes duration of mobile robots moving in periodic trajectories.  The dataset includes measurements of GNSS-RTK and two IMUs mounted on different wheels and one IMU mounted on the chassis of the robot. This dataset enables reproduction of our results and supports further research in inertial-based mobile robot localization. The dataset is publicly accessible at \href{https://github.com/ansfl/WMINet}{our GitHub repository}. 
\end{enumerate}
\noindent We evaluate our approach on real-world data collected from a mobile robot and demonstrate that it outperforms both the model-based and learning based baselines. We also demonstrate the effectiveness of the wheelbase constraint (WC) in improving the positioning performance.\\
\noindent The rest of the paper is organized as follows: Section
\ref{m-based} gives an overview of wheel-mounted inertial navigation and MoRPINet Section \ref{wnet_f} Presents our proposed approach, WMINet and the WC. Next, in Section \ref{dataset}, a detailed description of the mobile robot datasets is provided while in Section \ref{aandres} experimental results are given and analyzed. Lastly, Section \ref{conc} derives the conclusions of this research.
\section{Mobile Robot Inertial Navigation}
\label{m-based}
1) model-based wheel-mounted inertial navigation and 2) MoRPINet, a deep-learning approach for mobile robot positioning
\subsection{Wheel-Mounted Inertial Navigation}
\begin{figure}[h]
    \centering
    \includegraphics[width=0.7\linewidth]{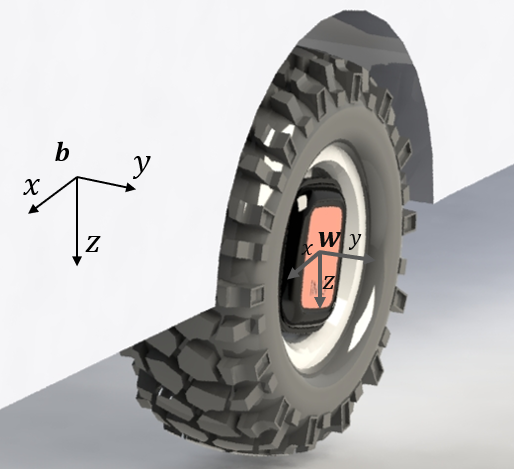}
    \caption{Axis definitions for the wheel-fixed frame (w-frame) and the body-fixed frame (b-frame)}
    \label{fig:wheel-trans}
\end{figure}

The equations of motion for an INS are typically formulated in the navigation frame, using north-east-down (NED) coordinates \cite{groves2015principles}. The rate of change of the position vector is given by:
\begin{equation}
\dot{\mathbf{p}}^n = \mathbf{v}^n  
\label{eq:pos_eq_1}
\end{equation}
\noindent where $\mathbf{p}^n$ is the position vector expressed in the n-frame, and $\mathbf{v}^n$ is the velocity vector expressed in the n-frame. The velocity rate of change is:
\begin{equation}
 \mathbf{\dot{v}^{n}}=\mathbf{T^{n}_{b}} \mathbf{f^{b}}+ \mathbf{{g}^n}  
\label{pos_eq}
\end{equation}
\noindent where $\mathbf{g}^n$ is the gravity vector expressed in the $n$-frame, while $\mathbf{T}_b^n$ is the transformation matrix that converts data from the body frame to the $n$-frame. The term $\mathbf{f}^b$ denotes the specific force vector as measured in the body frame.\\The transformation matrix rate of change is given by:
\begin{equation}
\dot{\mathbf{T}}_n^b = \mathbf{T}_n^b \boldsymbol{\Omega}^{b}
\label{eq:tran_eq}
\end{equation}
\noindent where $\boldsymbol{\Omega}^{b}$ is the skew-symmetric matrix of the angular rate measured by the gyroscope, expressed in the body frame. It is important to note that, given the use of low-cost inertial sensors and the short time durations considered in our scenarios, the effects of the Earth's rotation rate and transport rate are neglected in (\ref{pos_eq}) and (\ref{eq:tran_eq}).\\
In the context of mobile robot navigation, a local coordinate frame (b-frame)
is employed. This frame is defined at the robot’s initial position, with its axes aligned to the north-east-down directions.
\noindent Further, in a wheel-mounted configuration, compute the inertial navigation solution, it is necessary to transform the data collected from the wheel-mounted IMU into the b-frame. Figure \ref{fig:wheel-trans} illustrates the coordinate system of the wheel frame (denoted as \(w\)) in relation to the body frame (denoted as \(b\)).
\noindent The transformation involves calculating the phase angle of the wheel at each timestamp. As defined in \cite{collin2014mems}, the phase angle is:
\begin{equation}
\alpha(t)= \int_{t}\omega_z(\tau) d\tau
\label{eq:tran2}
\end{equation}
\noindent where \(\omega_z\) is the measured angular velocity and t is the time. \\
\noindent By using (\ref{eq:tran2}) we can calculate the transformation between the v-frame to the b-frame:
\begin{equation}
\boldsymbol{C_{b}^{w}} = \begin{bmatrix} 
cos(\alpha(t)) & sin(\alpha(t)) & 0 \\
-sin(\alpha(t)) & cos(\alpha(t)) & 0 \\
0 & 0 & 1 
\end{bmatrix}
\label{eq:tran1}
\end{equation}
\noindent Thus the angular velocity vector in the v-frame is
\begin{equation}
\boldsymbol{\omega^b}= \boldsymbol{C_{w}^{b}} \boldsymbol{\omega^w}
\label{eq:ang_vel}
\end{equation}
\noindent where \(\boldsymbol{\omega^w}\) is the measured angular velocity from the wheel-mounted IMU. \\
\noindent In the same manner, the specific force vector expressed in the v-frame is:
\begin{equation}
\boldsymbol{f^b}= \boldsymbol{C_{w}^{b}} \boldsymbol{f^w}
\label{eq:spec_vec}
\end{equation}
\noindent where \(\boldsymbol{f^w}\) is the measured specific force from the wheel-mounted IMU.
To compute the navigation solution, the angular velocity vector (\ref{eq:ang_vel}) and  the specific force vector (\ref{eq:spec_vec}) are substituted into (\ref{eq:pos_eq_1})-(\ref{eq:tran_eq}).
\subsection{MoRPINet Framework}
\begin{figure}[h]
    \centering
    \includegraphics[width=1.1\linewidth]{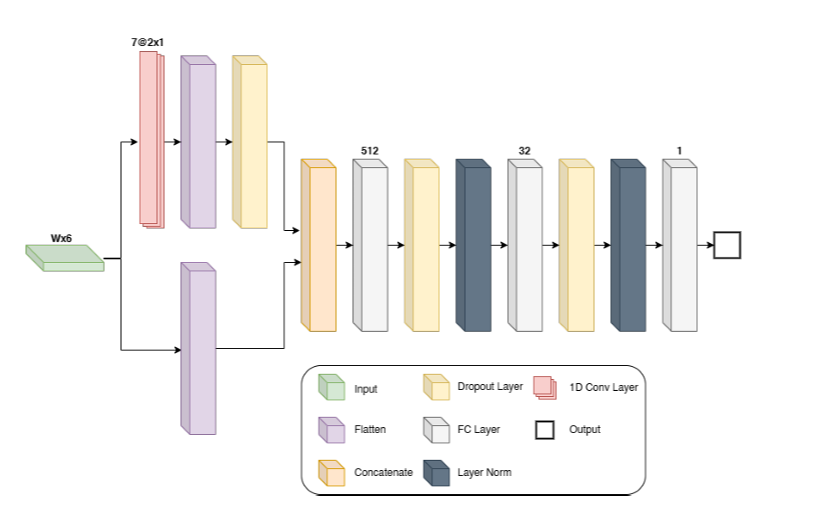}
    \caption{MoRPINet: inertial-based architecture to estimate the distance of a mobile robot \cite{etzion2024snakeinspiredmobilerobotpositioning}.}
    \label{fig:morpinet}
\end{figure}

The MoRPINet framework \cite{etzion2024snakeinspiredmobilerobotpositioning} is a deep neural network (DNN) designed to estimate the distance traveled by the robot, while the heading angle calculated by the Madgwick algorithm. The network structure consist of uses a 1D-CNN and fully concted (FC) layers as shown in Figure \ref{fig:morpinet}. 
\noindent The regressed distance and heading angle are then used to estimate the 2D position components of the platform:
\begin{equation}
\begin{split}
    x_{i+1} &= x_i + d_{\text{net},i} \cos {\psi}_i \\
    y_{i+1} &= y_i + d_{\text{net},i} \sin {\psi}_i
\end{split}
\label{morpinet_eq}
\end{equation}
\noindent where x and y are the platform position components, ${\psi}_i$ is the heading angle at index i, and ${d_{\text{net},i}}$ is the distance at index i.
\section{Wheel-Mounted Inertial Network}
\label{wnet_f}
\begin{figure*}[h]
    \centering
    \includegraphics[width=0.75\linewidth]{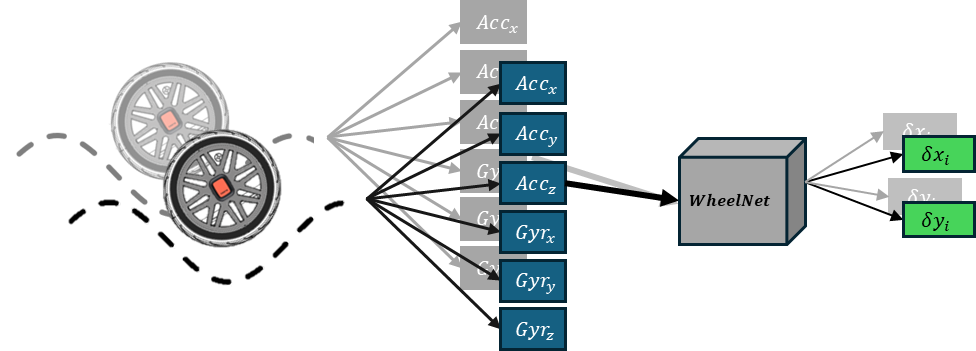}
    \caption{WMINet module estimates the mobile robot displacement based on wheel-mounted inertial sensor measurements. WMINet can be applied with one or two-wheel, two wheel-mounted  inertial sensors.}
    \label{fig:wheelscheme}
\end{figure*}

Mobile robots equipped with chassis-based inertial sensors often generate noisy inertial readings, characterized by a low signal-to-noise ratio, particularly when moving at a nearly constant velocity trajectories. The same phenomena occurs when using wheel-mounted sensors. 
To overcome this limitation we  employ periodic motion trajectories and design a learning algorithm for the robot positioning using only wheel-mounted IMU. The periodic motion is characterized by varying angular velocity and linear accelerations which in turn increases the inertial sensors  single to noise ratio.  Consequently, it provides distinct and robust features for neural networks, facilitating the extraction of accurate and relevant positioning information. To this end, we propose WMINet, a wheel-mounted neural inertial positioning approach for mobile robots moving in periodic motion. WMINet employs a specialized neural network architecture tailored for regression tasks. This architecture combines 2D convolutional layers (2D-CNN) to estimate the robot displacement, relying exclusively on the inertial sensor readings. Next, we extent WMINet for two wheel-mounted IMUs taking into account the wheelbase constraint. Our proposed approach is illustrated in Figure \ref{fig:wheelscheme}. 
\subsection{WMINet architecture}
The proposed network combinesof a multi-head architecture with 2D-CNN and FC layers to predict the robot displacement over time. Initially, raw IMU data is processed separately through a 2D-CNN layer to extract features while preserving temporal relationships. These features are concatenated pass through one more 2D-CNN, flattened, and passed through two FC layers (512 and 32 neurons). The output represents the displacement of the mobile robot. To correspond with the dimensions of the GNSS-RTK, the output is structured into five one-second intervals. To introduce non-linearities and enhance the model's ability to learn complex motion patterns, all convolutional and fully connected layers employ the rectified linear unit (ReLU) activation function \cite{agarap2018deep} as define in (\ref{relufunc}). 
\begin{equation}
    \text{ReLU(\textbf{x})} = \max(0, \textbf{x})
    \label{relufunc}
\end{equation}
\noindent where $\textbf{x}$ represent the output of the hidden layers.
\begin{figure}[h]
    \centering
    \includegraphics[width=1.0\linewidth]{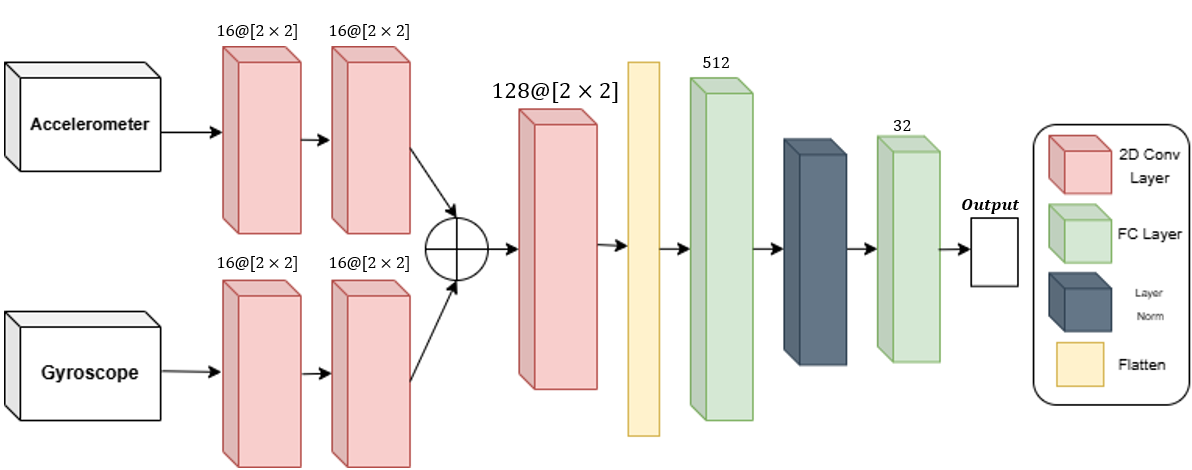}
    \caption{WMINet architecture for estimating the position of the mobile robot.}
    \label{fig:wheelarchitecture}
\end{figure}
\noindent Figure \ref{fig:wheelarchitecture} illustrates the architecture of our neural network. 
The numbers above each layer represents the  number of channels, while the values in brackets indicate the kernel size.
 \noindent Assuming a filter (or kernel) of size $m_1 \times m_2$, the resulting output of the convolutional layer can be expressed as follows \cite{goodfellow2016deep}:
\begin{equation}
\mathbf{C}_{ij}^{(\ell)} = \sum_{\alpha=0}^{m_1} \sum_{\beta=0}^{m_2} \bm{\omega}_{\alpha \beta}^{(r)} \bm{a}_{(i+\alpha)(j+\beta)}^{(\ell-1)} + \bm{b}^{(r)} 
\label{convfunc}
\end{equation}
\noindent where \( \bm{\omega}_{\alpha \beta}^{(r)} \) is the weight in the \( (\alpha, \beta) \) position of the \( r \)-th convolutional layer, \( \bm{b}^{(r)} \) represents the bias of the \( r \)-th convolutional layer, and \( \bm{a}_{ij}^{(\ell-1)} \) is the output of the preceding layer. \\
\noindent The fully-connected layers are built by a number of neurons. The following equation expresses the output of each neuron:
\begin{equation}
\bm{z}_i^{(\ell)} = \sum_{j=1}^{n_{\ell-1}} \bm{\omega}_{ij}^{(\ell)} \bm{a}_j^{(\ell-1)} + \bm{b}_i^{(\ell)}
\label{FC}
\end{equation}
\noindent where \( \bm{\omega}_{ij}^{(\ell)} \) is the weight of the \( i \)-th neuron in the \( \ell \)-th layer associated with the output of the \( j \)-th neuron in the \( (\ell-1) \)-th layer, \( \bm{a}_j^{(\ell-1)} \), \( \bm{b}_i^{(\ell)} \) represents the bias in layer \( \ell \) of the \( i \)-th neuron, and \( n_{\ell-1} \) represents the number of neurons in the \( \ell-1 \)-th layer.
\\
\noindent The input of the neural network is  sperately for accelerometer and gyroscope, each of them pass through convolution layer and ReLU activation function (\ref{relufunc}) as shown in (\ref{layer_1_acc}) and (\ref{layer_1_gyr}).
\begin{equation}
\mathbf{h_{1_{acc}}}=\text{ReLU}[\mathbf{C}_{ij}^{(0)}(\mathbf{X}_{acc})]
\label{layer_1_acc}
\end{equation}
\noindent where $\mathbf{X}_{acc}$ is the accelerometer input of the neural network.
\begin{equation}
\mathbf{h_{1_{gyro}}}=\text{ReLU}[\mathbf{C}_{ij}^{(0)}(\mathbf{X}_{gyro})]
\label{layer_1_gyr}
\end{equation}
\noindent where $\mathbf{X}_{gyro}$ is the gyroscope input of the neural network.\\
\noindent The neural network then applies an additional convolutional layer separately to the accelerometer and gyroscope data. The ReLU activation function is used in this step.
\begin{equation}
\mathbf{h_{2_{acc}}}=\text{ReLU}[\mathbf{C}_{ij}^{(1)}(\mathbf{h_{1_{acc}}})]
\label{layer_2_acc}
\end{equation}
\begin{equation}
\mathbf{h_{2_{gyro}}}=\text{ReLU}[\mathbf{C}_{ij}^{(1)}(\mathbf{h_{1_{gyro}}})]
\end{equation}
\noindent After applying two convolutional layers separately to the accelerometer and gyroscope data, we concatenate the resulting (\ref{conc_func}). This step allows the network to integrate information from both axis of the sensor and enabling it to extract shared patterns and relationships between the accelerometer and gyroscope features.
\begin{equation}
\mathbf{h_{concat}}=\text{concat}(\mathbf{h_{2_{gyro}}},\mathbf{h_{2_{acc}}})
\label{conc_func}
\end{equation}
\noindent After the convolutional layer, the concatenated output $h_{concat}$ is passed through another convolutional layer, followed by two fully connected layers:
\begin{equation}
\mathbf{\delta p} = \mathbf{h_{5}}(\mathbf{h_{4}}(\mathbf{C}_{ij}^{(2)}(\mathbf{h_{concat}} )))
\label{delta_p_eq}
\end{equation}
\noindent where $h_{4}$ and $h_{5}$ are two fully connected layers, and $\mathbf{\delta p}$ is the 2D mobile robot displacement $\mathbf{\delta p}$ = $[\delta x ,\delta y]^T$
where $\delta x$ is the displacement in the x component expressed in the navigation frame and  $\delta y$ is the displacement in the y component expressed in the navigation.
\subsection{Position Update}
To update the robot's position, we use the neural network output (\ref{delta_p_eq}) such that: 
\begin{equation}
    \begin{aligned}
        x_i &= x_{i-1} + \delta x_i \\
        y_i &= y_{i-1} + \delta y_i
    \end{aligned}
    \label{dis}
\end{equation}
\noindent where $x_i$ represent the $x$ coordinate position of the mobile-robot, $y_i$ represent the $y$ coordinate position of the mobile robot and i is the time index.
\noindent The position update frequency of the robot corresponds to the GNSS-RTK measurement rate, which operates at 5 Hz.
\noindent \subsection{Training process - WMINet} The training process aims to determine the weights and biases of all neurons that make up the DNN, to achieve the lowest possible loss function score. The loss function that used in this study is the mean squared error (MSE) loss function:
{
\begin{equation}
    \text{MSE} = 
    \frac{1}{n} \sum^{n}_{i=1}||\big(\bm{\delta \hat{p}}_i - \bm{\delta {p}}_i\big)^2||
    \label{lossfunc}
\end{equation}

}
\noindent where  $\bm \delta \hat p_i$ represents the ground truth (GT), $\bm \delta p_i$ represents the predicted values, which are the increments of distance and n represents the number of samples. 
\noindent To generate a prediction, the input passes through steps (\ref{convfunc})-(\ref{lossfunc}) or (\ref{twowheelloss}) if it is wheel tracks, in a process known as forward propagation \cite{hirasawa1996forward}. 
The learning process uses a stochastic gradient descent method, where the weights and biases are updated as follows:
\begin{equation}
\boldsymbol{\theta} = \boldsymbol{\theta} - \eta \nabla_{\theta} J(\boldsymbol{\theta}), \quad \text{where} \quad \boldsymbol{\theta} = 
\begin{bmatrix}
\boldsymbol{\omega} \\ 
\boldsymbol{b}
\end{bmatrix}^T
\end{equation}
\noindent In this context, $J(\boldsymbol{\theta})$ represents the loss function, with $\boldsymbol{\theta}$ being the vector of weights and biases. 
The term $\eta$ denotes the learning rate, while $\nabla_{\theta}$ is the gradient operator used to minimize the loss function .In the proposed method, the WMINet model utilized the Adaptive Moment Estimation (Adam) optimizer \cite{bock2018improvement} to enhance convergence efficiency. The training process was 400 epochs with a batch size of 128. An initial learning rate of 0.002 was used summary of the hyperparameters shown in Table \ref{hparam_wmin}.
\begin{table}[!h]
\centering
\caption{Hyperparameters parameters for WMINet.}
\resizebox{0.45\textwidth}{!}{ 
\begin{tabular}{|c|c|c|c|c|}
\hline
\textbf{Learning rate} & \textbf{Batch Size} & \textbf{Epoch} &\textbf{Window Size}  \\ \hline
0.002 & 128 & 400 & 120 \\ \hline
\end{tabular}}
\label{hparam_wmin}
\end{table}
\noindent \subsection{Wheelbase Constraint}
\begin{figure}[h]
    \centering
    \includegraphics[width=0.7\linewidth]{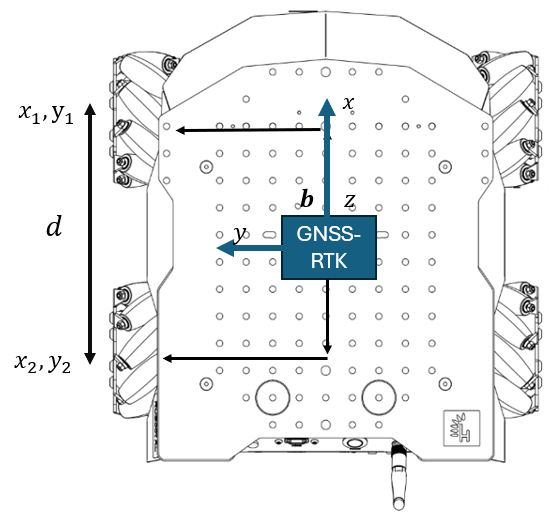}
    \caption{Scheme of translation between the GNSS-RTK to the each of wheel position.}
    \label{fig:diswheel}
\end{figure}

For the WC case, we use two IMUs and enforce a constraint on the wheel distance. Since the distance between the wheels remains constant, we first predict the displacement of each wheel and then translate the result to the wheel frame, as illustrated in Figure \ref{fig:diswheel}. The constraint between the wheels is then incorporated into the loss function, as shown in (\ref{losswtc}).
\begin{equation}
\text{\(L_d\)} = d - \sqrt{({x}_1 - {x}_2)^2 + ({y}_1 - {y}_2)^2} 
\label{losswtc}
\end{equation}
\noindent where \(x_i=x_i+\delta x_i\) and  \({y}_i=y_i+\delta y_i\) where \(\delta x_i\) and \(\delta y_i\) represent the predictions of the neural network, $x_i$ and $y_i$ represent the distance translation for each of the wheels and \( \mathrm{d} \) represent the constant distance between the wheels.
\noindent We can define the loss function for each of the wheel and the loss function for the distance of wheels by using (\ref{lossfunc}):
\begin{equation}
    \text{MSE}_{1}=\text{MSE}(\delta x_1, \delta y_1)
\end{equation}
\begin{equation}
    \text{MSE}_{2}=\text{MSE}(\delta x_2, \delta y_2)
\end{equation}
\noindent The total loss is expressed using the parameters \(\alpha\), \(\beta\), and \(\gamma\), which correspond to different components of the loss function. These parameters were determined through optimization on our dataset. The final formulation of the total loss is:
\begin{equation}
J = \alpha \text{MSE}_1 + \beta \text{MSE}_2 + \gamma L_d    
\label{twowheelloss}
\end{equation}
\noindent where the value of \(\alpha\) is 0.5 the value of \(\beta\) is 0.4 and the value of \(\gamma\) is 0.1. $\text{MSE}_1$ is the loss between the front wheel to the GT and  $\text{MSE}_2$ is the loss between the rear wheel to the GT.
\section{Dataset Generation}
\label{dataset}
\subsection{ Experiment Setup}
\begin{figure}[!h]
    \centering
    \includegraphics[width=1.0\linewidth]{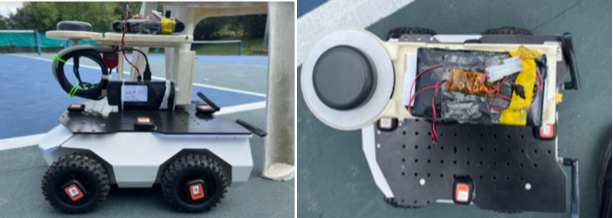}
    \caption{Right: The MRU-P on top of the ROSbot-XL with several Dots, Left: The Dot IMUs mounted on the wheels of the ROSbot-XL.}
    \label{fig:ROSBOTxl}
\end{figure}

A ROSbot-XL \cite{rosbotxl} mobile robot equipped with Movella Xsense Dots IMU \cite{movella_dot} on each wheel, was used to recored our dataset. The DOT software allows synchronization between the IMUs, The associated noise and bias values of the accelerometer and gyroscope are presented in Table~\ref{tab:gyro_accel}. Both sensors provide measurements at the rate of 120Hz. The ROSbot-XL was also equipped with an MRU-P \cite{inertial_labs} with RTK-GNSS capabilities to provide the ground truth (GT) position at a sampling rate of 5Hz. The ROSbot-XL equipped with our sensors and supporting electronics is shown in Figure \ref{fig:ROSBOTxl}. Controlled by an RC controller ,ROSbot-XL has dimensions of 332 x 325 x 133.5 mm and a wheel diameter of 100 mm.
\begin{table}[h!]
\centering
\small 
\caption{Xsense DOT IMU specifications \cite{movella_dot}.}
\begin{tabular}{|c|c|c|}
\hline
                & \textbf{Gyro} & \textbf{Accelerometer} \\ \hline
\textbf{Bias}    & $10 \, [^\circ/\text{h}]$   & $0.03 \, [\text{mg}]$         \\ \hline
\textbf{Noise}   & $0.007 \, [^\circ/\text{s}/\sqrt{\text{Hz}}]$ & $120 \, [\mu g/\sqrt{\text{Hz}}]$ \\ \hline
\end{tabular}
\label{tab:gyro_accel}
\end{table}
\noindent \subsection{Dataset} 
\begin{figure}[h]
    \centering
    \includegraphics[width=1.0\linewidth]{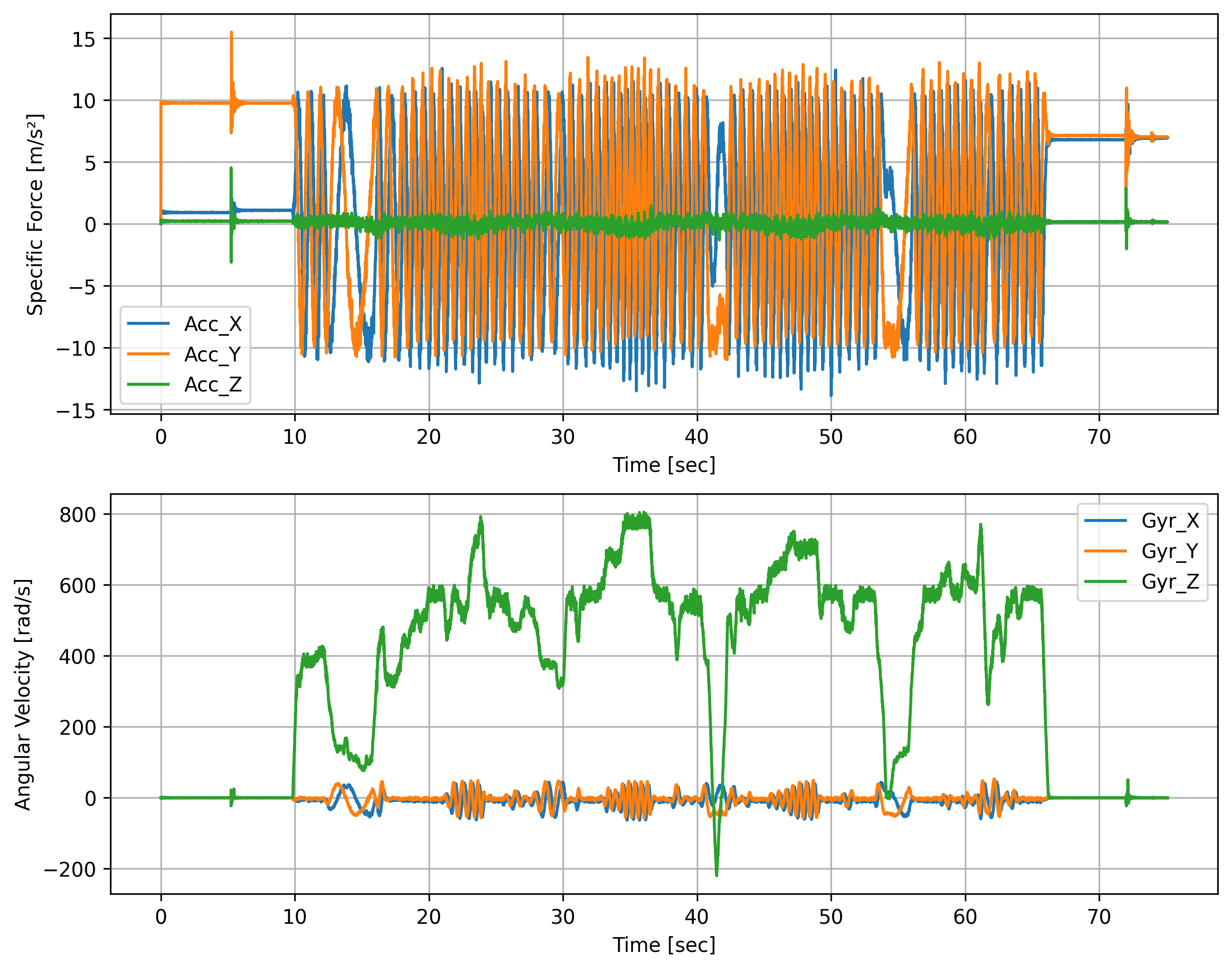}
    \caption{Wheel-mounted inertial measurements reading of trial-01 for a wheel-mounted IMU during periodic motion.}
    \label{imu_readings}
\end{figure}

Twenty-six periodic motion trajectories and two straight-line trajectories were recorded during field experiments with a total time of 38 minutes for a single IMU and 190 minutes from the five IMUs (four on the wheels and one on the chassis). Each recording contains the raw inertial measurements and associated GT position. An example of accelerometer and gyroscope reading during a periodic trajectory is presented in Figure \ref{imu_readings}.
Each recording session began with 10 seconds in stationary conditions which was utilized for synchronize between the Dot IMUs and the GNSS-RTK. For the same reason, the end of the trajectory the robot was placed in stationary conditions for a duration of 10 seconds. 
\subsubsection{Train Dataset}
The neural network training dataset consists of twenty four trajectories, all exhibiting periodoc  motion. The recording duration for each trajectory ranges from 2 to 5 minutes, resulting in a total of 32 minutes of data per IMU and 128 minutes for the entire training dataset.
The IMU data was segmented into time windows, each paired with corresponding GT values. The targets were derived from pairs of GNSS-RTK measurements, resulting in the displacement $\delta x$  and $\delta y$.
The IMU data was segmented based on the intervals between consecutive RTK measurements. Since the IMU operates at a frequency of 120 Hz and the RTK at 5 Hz, each RTK sample corresponds to twenty four IMU samples. We used a window size of 120 IMU samples,corresponding to one second windows. 
\noindent \subsubsection{Test Dataset}
The test dataset consists of two trajectories, each covering a fixed distance of approximately 12 meters, recorded while the robot performed a periodic motion trajectory, each trajectory is between 2 to 4 minutes, resulting in a total of 7 minutes of data per IMU and 28 minutes for the entire testing dataset.
\section{Analysis and Results}
\label{aandres}
This section describe the evaluation metrics used to measure the performance of our proposed method and presents the experimental results. 
\subsection{Performance Metrics}
The chosen metrics were specifically selected to align with the objectives of evaluating the position error. The first metric is the position root mean square error (PRMSE):
\begin{equation}
\text{PRMSE}(\boldsymbol{x_i}, \boldsymbol{\hat{x}_i}) = \sqrt{\frac{\sum_{i=1}^{N} |\boldsymbol{x_i} - \boldsymbol{\hat{x}_i}|^2}{N}}
\label{prmse}
\end{equation}
\noindent where $\boldsymbol{x_i}$ is the position vector of the ground truth and $\boldsymbol{\hat{x}_i}$ is the predicated position vector after we sum each $\delta x_i$ and $\delta y_i$ of the predict trajectory.\\
The second metirc is the total distance error (TDE):
\begin{equation}
\text{TDE} (\%) = \frac{\text{{PRMSE}}}{\text{{D}}} \times 100
\label{tde}
\end{equation}
\noindent where D is the length of the periodic motion trajectory.
\subsection{Results}
To evaluate WMINet performance we used the two trajectories of the test dataset. Those were also used with the standard wheel-mounted inertial navigation (WMIN) approach (\ref{eq:pos_eq_1})-(\ref{eq:spec_vec}) and MoRPINet (\ref{morpinet_eq}). To mitigate the model-based approach error, we applied it assuming 2D motion ($f_z$ = $\omega_x$ = $\omega_y$ = 0). Also, the MoRPINet approach was applied to the chassis installed IMU to insure a fair comparison as this approach was originally developed for non-wheel mounted IMU.
\\
\noindent Table 3 gives the results of all three approaches in terms of PRMSE and TDE. As expected, there was a significant error using the WMIN approach with an average PRMSE of 177.49 [m]. The MoRPINet approach solution, based on the regressed distance from the network and heading angle from the Madgwick algorithm, obtained an average PRMSE of 3.45 [m]. With our WMINet solution, we achieved the lowest PRMSE of 1.54 [m], improving MoRPINet by 55\%.
\\
\noindent In addition we also present the results in terms of TDE. The WMIN approach present a significant error, with an average TDE of 99.0\%. The MoRPINet approach, reduced the TDE, achieving an average of 31.60\%. Our proposed WMINet solution further improved accuracy, achieving the lowest TDE of 12.91\%.
\begin{table}[h!]
\centering
\caption{Test dataset results for the model and learning-based baselines and our WMINet approach.}
\resizebox{0.5\textwidth}{!}{ 
\begin{tabular}{|c|c|c|c|}
\hline
\textbf{Method} & \textbf{Trajectory} & \textbf{PRMSE [m]} & \textbf{TDE [\%]} \\ \hline
\multirow{3}{*}{WMIN (model-based)} & Traj. 1 & 186.63 & 99.00 \\ \cline{2-4}
                               & Traj. 2 & 168.35 & 99.00 \\ \cline{2-4}
                               & \textbf{Average} & 177.49 & 99.00 \\ \hline
\multirow{3}{*}{MoRPINet (baseline)} & Traj. 1 & 3.70 & 32.04 \\ \cline{2-4}
                           & Traj. 2 & 3.20 & 31.16 \\ \cline{2-4}
                           & \textbf{Average} & 3.45 & 31.60 \\ \hline
\multirow{3}{*}{WMINet (ours)} & Traj. 1 & 1.37 & \textbf{11.50} \\ \cline{2-4}
                                    & Traj. 2 & 1.71 & \textbf{14.15} \\ \cline{2-4}
                                    & \textbf{Average} & 1.54 & \textbf{12.91} \\ \hline
\end{tabular}
}
\label{combined_results}
\end{table}
\subsubsection{Wheelbase Constraint}
For the wheelbase constraint the same hyperparameters of WMINet was used only with the new loss function (\ref{twowheelloss}).
\begin{table}[h!]
\centering
\caption{Test dataset results for our WMINet with and without the WC constraint.}
\resizebox{0.5\textwidth}{!}{ 

\begin{tabular}{|c|cl|clc|}
\hline
\textbf{Method}  & \multicolumn{2}{c|}{\textbf{WMINet}} & \multicolumn{3}{c|}{\textbf{WMINet with WC}}                   \\ \hline
                 & \multicolumn{2}{c|}{PRMSE {[}m{]}}   & \multicolumn{2}{c|}{PRMSE {[}m{]}} & Improvment {[}\%{]}       \\ \hline
Traj.1           & \multicolumn{2}{c|}{1.37}            & \multicolumn{2}{c|}{0.94}          &    {31.38}                       \\ \hline
Traj.2           & \multicolumn{2}{c|}{1.71}            & \multicolumn{2}{c|}{1.35}          &    {21.02}                       \\ \hline
\textbf{Average} & \multicolumn{2}{c|}{\textbf{1.54}}   & \multicolumn{2}{c|}{\textbf{1.16}} & \textbf{24.0}               \\ \hline
\end{tabular}}
\label{wminetwc}
\end{table}
%
Table \ref{wminetwc} gives the results of the new approach of WMINet with WC.  The WNINet approach solution obtained an average PRMSE of 1.54 [m], while with WMINet with WC we achieve a PRMSE of 1.16 [m]. Compared to our WMINet, WMINet with WC improves the results by 24\%.     
\begin{figure}[!h]
    \centering
    \includegraphics[width=1.0\linewidth]{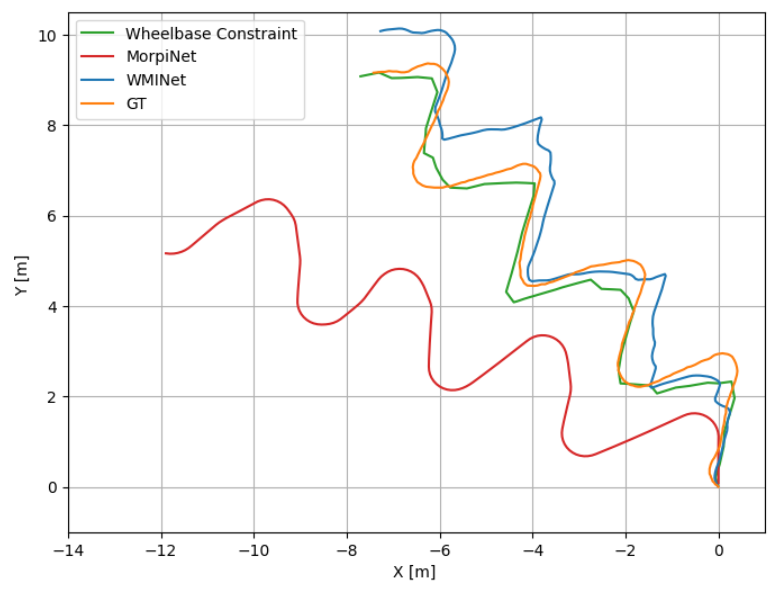}

    \caption{Reconsecrated trajectories of our proposed approach and the associated GT trajectory.}
    \label{res_traj}
\end{figure}
\noindent Figure \ref{res_traj} presents the trajectory of the mobile robot for each method except for the WMIN approach due to its rapidly increasing position error. While MoRPINet captures the periodic motion effectively it suffers from error diverging the trajectory from the GT. While, WMINet does not replicate a perfect periodic trajectory as closely, it provides a more accurate estimation and maintaining the direction of the GT trajectory. With the wheelbase constraint the error reduces resulting in an accurate prediction of the GT trajectory.
\section {Conclusions}
\noindent In many real-world scenarios, mobile robots must rely on inertial sensors for positioning, but these sensors' errors cause position solution drift. To address this issue, we propose WMINet, an end-to-end neural network designed to estimate the robot’s position using wheel-mounted inertial sensor readings. To this end, the robot must follow a periodic motion trajectory to enrich the inertial readings and allow accurate regression by the network. In order to further improve performance, we implemented a wheelbase constraint since the distance between the wheels is known in advance. To this end, we derived a dedicated loss function taking this information into account. A dataset of 38 minutes of inertial recordings was collected for training and testing. Our approach was compared to a model-based approach and a learning-based approach. Our analysis of WMINet provided an improvement of more than 55\% over the other two approaches. With the addition of the WC, the improvement rate increased to 66\%. \\
\noindent In summary, WMINet provides a precise positioning solution for mobile robots in real-world scenarios where only inertial sensors are available. These situations frequently occur when GNSS is not available for robots that operate outdoors or in situations of high maneuverability and poor lighting conditions while working indoors. As a consequence, our approach enables navigation in GNSS-denied environments and improves robustness in challenging conditions by using neural networks and incorporating physical constraints. Thus, WMINet bridges the pure inertial gap and enables seamless robot navigation using only inertial sensors.
\label{conc}
\section{Acknowledgments}
G.V. was supported by the Maurice Hatter Foundation.
\bibliographystyle{ieeetr}
\bibliography{bibilography}
%
\end{document}